\documentclass{article}

 \usepackage[preprint]{neurips_2025}

\usepackage[utf8]{inputenc} 
\usepackage[T1]{fontenc}    
\usepackage{hyperref}       
\usepackage{url}            
\usepackage{booktabs}       
\usepackage{amsfonts}       
\usepackage{nicefrac}       
\usepackage{microtype}      
\usepackage{xcolor}         
\usepackage{amsmath} 
\usepackage{enumitem} 

\usepackage{graphicx} 
\usepackage{booktabs} 
\usepackage{multirow} 

\title{Permutation-Invariant Transformer Neural Architectures for Set-Based Indoor Localization Using Learned RSSI Embeddings}

\author{Anonymous Authors}

\author{%
  Aris J.~Aristorenas\\
  Department of Computer Science\\
  Stanford University\\
  Stanford, CA 94305 \\
  \texttt{aaris824@stanford.edu} \\
}

\begin{document}

\maketitle

\begin{abstract}
 We propose a permutation-invariant neural architecture for indoor localization using RSSI scans from Wi-Fi access points. Each scan is modeled as an unordered set of (BSSID, RSSI) pairs, where BSSIDs are mapped to learned embeddings and concatenated with signal strength. These are processed by a Set Transformer, enabling the model to handle variable-length, sparse inputs while learning attention-based representations over access point relationships. We evaluate the model on a dataset collected across a campus environment consisting of six buildings. Results show that the model accurately recovers fine-grained spatial structure and maintains performance across physically distinct domains. In our experiments, a simple LSTM consistently outperformed all other models, achieving the lowest mean localization error across three tasks (E1 - E3), with average errors as low as 2.23 m. The Set Transformer performed competitively, ranking second in every experiment and outperforming the MLP, RNN, and basic attention models, particularly in scenarios involving multiple buildings (E2) and multiple floors (E3). Performance degraded most in E2, where signal conditions varied substantially across buildings, highlighting the importance of architectural robustness to domain diversity. This work demonstrates that set-based neural models are a natural fit for signal-based localization, offering a principled approach to handling sparse, unordered inputs in real-world positioning tasks.
\end{abstract}

\section{Introduction}

Indoor localization remains an open problem in the deployment of real-world navigation systems, with applications ranging from asset tracking \cite{lee2019bluetooth,kao2017hybrid,hayward2022survey}, and assistive technologies in robotics \cite{huang2023indoor,nazemzadeh2015indoor}. Unlike outdoor environments, which benefit from satellite-based positioning systems such as GPS, indoor settings pose unique challenges due to signal attenuation, multipath effects, and structural occlusions. As a result, researchers have increasingly turned to alternative modalities in IoT (Internet-of-Things). These include low-cost, and widely available sensors such as Radio-Frequency Identification Tags (RFID) \cite{ting2011study,zhu2019review}, Wi-Fi access points (see Section ~\ref{sec:related-work-trad-dl}), Bluetooth Low Energy (BLE) beacons \cite{jianyong2014rssi,bai2020low}, and Ultra-wideband tags (UWB) \cite{alarifi2016ultra,mazhar2017precise}.

In recent years, deep learning models have emerged as promising candidates for learning signal-to-position mappings directly from data, sidestepping the need for explicit signal propagation models or hand-crafted feature engineering. However, most prior work either focuses on small-scale datasets or assumes access to complete knowledge of the environment (such as the positions of access points), limiting generalizability. Furthermore, traditional architectures often treat signal data as ordered vectors, ignoring the inherent permutation invariance of signal sets where the order of access points is arbitrary and variable across signal scans.

In this work, we evaluate the performance of permutation-invariant neural architectures, specifically the Set Transformer, on the task of RSSI-based indoor localization purely from abundant Wi-Fi access points. We compare its effectiveness against standard neural baselines including the Multi-layer Perceptron (MLP), a vanilla Recurrent Neural Network (RNN), the Long-Short Term Memory (LSTM) model, and basic Attention models. Our evaluation spans multiple localization tasks in a real world university campus setting, which includes single-building single-floor scenarios, cross-building supervised learning across first floors of multiple buildings, and multi-floor prediction within a single building. The results provide insight into the tradeoffs between model complexity, generalization on a held out test set, and spatial reasoning in constrained indoor environments.

Our contributions are as follows: 1) We design a structured evaluation framework for indoor localization across multiple buildings and floors, using real-world signal data collected under a controlled protocol, 2) We present (to the best of our knowledge) the first application of Set Transformers to signal-based indoor positioning, treating each scan as an unordered set and learning signal-specific embeddings without requiring environment-specific engineering, 3) We conduct a comparative study against standard neural baselines (MLP, RNN, LSTM, Attention, Transformer), demonstrating that attention-based and permutation-invariant models can match or exceed traditional architectures across challenging localization tasks.

\section{Related Work}

We start by reviewing earlier works that addressed indoor localization using traditional neural network based models. Then we review set-based neural architectures.

\subsection{Indoor Localization Using Traditional Deep Learning}
\label{sec:related-work-trad-dl}

Numerous studies have applied standard deep learning models to RSSI-based indoor positioning, demonstrating improved accuracy over classical techniques.

Feedforward artificial neural networks (ANNs) or Multi-Layer Perceptrons (MLPs) have been shown to learn nonlinear mappings from RSSI to position coordinates \cite{alhomayani2020deep, chin2020intelligent}. Convolutional Neural Networks (CNNs) have also been adapted for localization by treating RSSI vectors or signal maps as spatially structured inputs, with attention-enhanced CNNs further improving performance \cite{qin2021ccpos, abid2021improved, mazlan2022fast, ibrahim2018cnn}. In addition, fingerprint-image-based CNN methods have demonstrated sub-meter accuracy using hierarchical learning \cite{shao2018indoor}.

Several works explore temporal dynamics in RSSI signals using Recurrent Neural Networks (RNNs) and Long-Short Term Memory (LSTM), which model time-sequential signal patterns for improved robustness in dynamic environments \cite{lukito2017recurrent, bai2019dl, hoang2019recurrent, sahar2018lstm}. Schmidt et al. introduced SDR-Fi, a deep learning system using channel state information (CSI) collected via software-defined radio, outperforming conventional RSSI systems \cite{schmidt2019sdr}. Ayyalasomayajula et al. proposed DLoc, integrating deep learning with autonomous map construction for robust, scalable localization \cite{ayyalasomayajula2020deep}.

These models typically assume fixed-length inputs and do not treat RSSI scans as unordered sets. In contrast, our work applies permutation-invariant architectures to the localization task, preserving the unordered nature of signal scans and learning generalized representations across increasing environment complexities (multiple buildings on a campus environment with multiple floors).

\subsection{Set-based Neural Architectures}

Set Transformer \cite{lee2019set} extends Deep Sets \cite{zaheer2017deep} by replacing simple pooling with attention-based mechanisms, enabling richer permutation-invariant representations via self-attention and pooling by multihead attention (PMA). We build on this framework by applying it to signal-based indoor localization, where unordered RSSI scans naturally align with the set input paradigm.

Before the introduction of permutation-invariant architectures, \cite{vinyals2015order} proposed handling sets using sequence models trained over multiple permutations. Unlike this approximate strategy, Deep Sets and Set Transformer incorporate invariance directly into the model structure through summation and attention.

The Perceiver \cite{jaegle2021perceiver} generalizes attention-based models to handle large, unordered inputs using iterative cross-attention to a fixed latent array. While not strictly permutation-invariant due to its use of positional encodings, it demonstrates how set-like architectures can be scaled to high-bandwidth perceptual inputs. Our approach differs in focusing on learned embeddings for BSSID-based (Basic Set Service Identifier) inputs and explicitly preserving permutation invariance for signal localization.

Other set-based architectures include Janossy pooling \cite{murphy2018janossy}, SetVAE \cite{kim2021setvae}, Slot Attention \cite{locatello2020object}, FSPool \cite{zhang2019fspool}, and SeTformer \cite{shamsolmoali2024setformer}. While these models address tasks such as generative modeling, object-centric reasoning, or adaptive pooling, our work applies permutation-invariant architectures specifically to signal-based localization with learned (RSSI, BSSID) embeddings.

To the best of our knowledge, Set Transformers have not been applied to RSSI-based indoor localization. Prior transformer-based approaches in this space typically rely on fixed-length inputs or impose sequence order \cite{wang2015deepfi, song2019cnnloc, hoang2020semi, savin2023exploring, nasir2024hytra}. In contrast, our work treats RSSI scans as permutation-invariant sets and evaluates generalization across indoor environments of increasing complexity.

\section{Problem Formulation}

We consider the problem of indoor localization based on wireless signal measurements of Received Signal Strength Indicator (RSSI) from Wi-Fi Access Points (AP). Specifically, given a single RSSI scan collected from a device capable of receiving RSSI (such as a mobile device, or laptop), the goal is to predict the user’s spatial location in 2-dimensional space. Each scan is modeled as an unordered set of Wi-Fi detections:
\begin{align}
S = \{(b_1, r_1), (b_2, r_2), \ldots, (b_n, r_n)\}
\end{align}

where $b_i \in \mathcal{B}$ is the BSSID (the MAC address) of a detected access point and $r_i \in \mathbb{R}$ is the corresponding RSSI value in units of dBm. The number of elements $n$ varies between scans due to environmental signal occlusion and device limitations.

Now let $f_\theta: \mathcal{S} \rightarrow \mathbb{R}^2$ denote a localization function parameterized by $\theta$, which maps an input scan $S$ to a 2D position estimate $(x, y)$. The function $f_\theta$ must satisfy two critical properties:

\begin{enumerate}
    \item \textbf{Permutation Invariance:} The ordering of elements in $S$ must not affect the output.
    \item \textbf{Set Generalization:} The model must handle variable-sized inputs and operate on previously unseen BSSIDs during inference.
\end{enumerate}

Permutation invariance is necessary because (BSSID, RSSI) pairs can arrive at the device in arbitrary order across scans. Set generalization is required because some access points may appear or disappear across scan windows, and the set of observed BSSIDs may differ at inference time.

To accommodate this, we treat scans as sets and build a model architecture that respects the inductive bias of set-structured data. Our learning objective minimizes the average Euclidean distance between predicted and true positions over a dataset of $N$ labeled scans:
\begin{align}
\mathcal{L}(\theta) = \frac{1}{N} \sum_{i=1}^N \left\| f_\theta(S_i) - (x_i, y_i) \right\|_2^2
\end{align}

To encode this inductive bias, we design $f_\theta$ as a permutation-invariant neural architecture based on the Set Transformer, detailed in Section~\ref{sec:methods}. This choice allows the model to process variable-length RSSI scans as unordered sets and to learn attention-based representations over access point interactions, even under signal sparsity or occlusion. We extend this formulation to real-world deployment scenarios by evaluating performance across multiple indoor configurations ~\ref{sec:data-collection}.

Lastly, note that in settings with multiple floors, we also consider an alternative formulation where the model predicts a 2D position $(x, y)$ along with a discrete floor label $z$, treated as a classification task. This multi-task learning extension is detailed in Section~\ref{sec:multi-task}.

\section{Methods}
\label{sec:methods}

We now outline the methods of this work starting with the input encoding, to the baselines considered, and the Set Transformer formulation applied to RSSI data.

\subsection{RSSI Embedding and Input Encoding}

Each RSSI scan is represented as an unordered set of detections from nearby Wi-Fi Access Points, where each element is a tuple $(b_i, r_i)$ consisting of a unique BSSID $b_i$ (a MAC address) and a received signal strength indicator $r_i \in \mathbb{R}$ in dBm. As previously mentioned, due to environmental occlusion and signal instability, the number of detections varies between scans.

The embedding layer enables the model to convert categorical BSSIDs into dense vectors that are learned during training. This allows the model to reason over structural patterns in RSSI sets, rather than raw identifiers. These embeddings capture contextual relationships between access points, and are essential for supporting permutation-invariant set processing.

Unlike raw integer IDs, which impose arbitrary ordinal relationships, or one-hot encodings, which treat all APs as orthogonal and unrelated, learnable embeddings provide a dense, continuous representation that is both scalable and expressive. This approach allows the model to discover latent structure across access points—such as spatial proximity, co-occurrence patterns, or coverage similarity—purely from training data. It also enables generalization to unseen BSSIDs, since the model can fall back to randomly initialized embeddings at test time, without requiring retraining or explicit reindexing.

To transform each scan into a suitable input for our permutation-invariant model, we apply the following encoding procedure:

\begin{enumerate}[label=(\roman*)]
\item \textbf{BSSID Embedding:} Each $b_i$ is mapped to a fixed-length learnable embedding vector $e_i \in \mathbb{R}^d$. The embedding table is initialized randomly and trained jointly with the model. Embeddings are updated via backpropagation only if the corresponding BSSID is observed during training.

\item \textbf{RSSI Concatenation:} The scalar $r_i$ is concatenated to its corresponding embedding $e_i$ to form an input vector $s_i = [e_i | r_i] \in \mathbb{R}^{d+1}$.

\item \textbf{Set Formation:} The resulting scan is encoded as a set $S = \{s_1, s_2, \dots, s_n\}$, where each $s_i \in \mathbb{R}^{d+1}$.

\item \textbf{Model Input:} The set $S$ serves as the input to the localization model $f_\theta: \mathcal{S} \rightarrow \mathbb{R}^2$, detailed in Section~\ref{sec:set-transformer}.
\end{enumerate}

To accommodate variable-length scans, each RSSI set is processed individually (batch size = 1), avoiding the need for explicit padding or masking. While this prevents efficient batching, it preserves the full fidelity of the unordered scan data. In future work, we plan to extend the model to support attention masking and larger batch sizes.

\subsection{Handling Variable-Length and Missing Inputs}

RSSI scans vary in cardinality due to environmental factors (such as complicated building infrastructure), differences in device sensitivity, and transient signal dropout. Therefore, the input set $S = \{s_1, s_2, \dots, s_n\}$ can vary in size across scans. Rather than forcing a fixed-length representation, we process each scan independently, maintaining the natural set structure of the data.

During training and inference, we set the batch size to one, allowing the model to process each scan as a variable-length set without the need for padding or masking. This design choice simplifies handling of missing access points and avoids introducing artifacts through arbitrary truncation or zero-padding. Although this limits batching efficiency, it preserves the integrity of permutation-invariant set processing.

Missing access points such as BSSIDs that are not detected in a given scan, are excluded from the input set. Let $\mathcal{B}$ denote the set of all unique BSSIDs observed across the dataset, and let $\mathcal{B}_{\text{scan}} \subset \mathcal{B}$ denote the subset detected in a particular scan. The input set for that scan is then defined as:
\begin{align}
S = \left\{ s_i = [e_i \,\Vert\, r_i] \;\middle|\; b_i \in \mathcal{B}_{\text{scan}} \right\}
\end{align}

No imputation or placeholder values are required for the undetected BSSIDs in $\mathcal{B} \setminus \mathcal{B}_{\text{scan}}$. The model is trained to infer position from the available subset of signals, relying on the joint structure encoded in $s_i$ vectors derived from observed RSSI values and BSSID embeddings. This design choice is critical for real-world deployment, where signal sparsity and dynamic network conditions are unavoidable.

\subsection{Multilayer Perceptron (MLP) Baseline}
\label{sec:mlp}

As a non-sequential, order-agnostic baseline, we implement a fully connected multilayer perceptron (MLP) to map fixed-length RSSI vectors to 2D spatial coordinates. Each RSSI scan is represented as a dense vector $x \in \mathbb{R}^d$, where each component corresponds to the signal strength from a unique BSSID observed in the training set. If a particular BSSID is not detected in a given scan, the corresponding value is set to $-100$ dBm, representing a signal dropout.

The MLP consists of three fully connected layers with ReLU activations. Let $W_1, W_2, W_3$ and $b_1, b_2, b_3$ denote the weight matrices and bias vectors of each layer. The forward pass of the model is given by:
\begin{equation}
\begin{aligned}
h_1 &= \text{ReLU}(W_1 x + b_1) \\
h_2 &= \text{ReLU}(W_2 h_1 + b_2) \\
\hat{y} &= W_3 h_2 + b_3
\end{aligned}
\end{equation}

where $x \in \mathbb{R}^d$ is the input RSSI vector and $\hat{y} \in \mathbb{R}^2$ denotes the predicted normalized coordinates.

To ensure consistent feature alignment across scans, the input vector $x$ follows a fixed BSSID ordering derived from training data. This fixed-length representation enables efficient batch training but lacks the robustness of set-based or sequence models to missing or reordered detections.

The MLP is trained using the same coordinate normalization and mean squared error objective described in Section~\ref{sec:training-objective}, providing a simple yet informative baseline for evaluating the effectiveness of more structured architectures.

\subsection{Recurrent Neural Network (RNN) Baseline}
\label{sec:rnn}

As an alternative to fixed-length vector models, we implement a recurrent neural network (RNN) baseline that treats each RSSI scan as a variable-length sequence of $(\text{BSSID}, \text{RSSI})$ pairs. This approach preserves the sequential structure of the input by sorting detected access points by descending signal strength and embedding each BSSID into a learnable vector space.

Let each scan be represented as a sequence $[(b_1, r_1), (b_2, r_2), \dots, (b_T, r_T)]$, where $T$ is the number of access points in the scan. For each pair, the BSSID $b_i$ is mapped to a learnable embedding vector $e_i \in \mathbb{R}^d$, and the RSSI value $r_i \in \mathbb{R}$ is concatenated to the embedding, forming $s_i = [e_i \,\Vert\, r_i] \in \mathbb{R}^{d+1}$.

The resulting sequence $[s_1, s_2, \dots, s_T]$ is passed to a vanilla RNN encoder, which processes one vector at a time and updates a hidden state $h_i$ according to:
\begin{align}
h_i = \tanh(W_h s_i + U_h h_{i-1} + b_h)
\end{align}

We take the final hidden state $h_T \in \mathbb{R}^{h}$ as a global representation of the scan and pass it through a fully connected layer to regress the normalized 2D position:
\begin{align}
\hat{y} = W_o h_T + b_o
\end{align}

The model is trained to minimize the mean squared error (MSE) between the predicted and true normalized coordinates, using the same loss function and normalization procedure as described in Section~\ref{sec:training-objective}. The batch size is set to 1 to allow handling of variable-length input sequences without padding or masking.

This baseline serves to evaluate whether simple sequence models can learn meaningful localization signals from ordered RSSI inputs, even without explicit attention or complex recurrence.

\subsection{Long Short-Term Memory (LSTM) Baseline}
\label{sec:lstm}

We extend the recurrent modeling approach by considering a Long Short-Term Memory (LSTM) network, which addresses the vanishing gradient problem and enables the model to capture longer-range dependencies across input sequences. Each RSSI scan is first sorted by descending signal strength and encoded as a sequence of vectors:
\begin{align}
s_i = [e_i \,\Vert\, r_i] \in \mathbb{R}^{d+1}, \quad i = 1, \dots, T
\end{align}

where $e_i$ is a learned embedding of BSSID $b_i$, and $r_i$ is the associated RSSI value. This produces a variable-length input sequence $S = [s_1, s_2, \dots, s_T]$. The sequence is passed through an LSTM encoder:
\begin{align}
h_i, c_i = \text{LSTM}(s_i, (h_{i-1}, c_{i-1}))
\end{align}
where $h_i \in \mathbb{R}^h$ is the hidden state and $c_i$ is the cell state at step $i$. We use the final hidden state $h_T$ as the global representation of the scan. This representation is passed through a feedforward head to regress the normalized 2D coordinates:
\begin{align}
\hat{y} = W_o h_T + b_o
\end{align}

The model is trained using the same coordinate normalization and mean squared error loss described in Section~\ref{sec:training-objective}. Like the RNN baseline, the LSTM is trained with batch size 1 to support variable-length sequences without padding.

Compared to vanilla RNNs, LSTMs provide improved stability and learning capacity on longer or more variable input sequences. This baseline evaluates whether such sequence-aware models benefit RSSI-based localization when ordered $(\text{BSSID}, \text{RSSI})$ sequences are used as input.

\subsection{Attention Mechanism Baseline}
\label{sec:attention}

We implement a simple attention-based model that learns to weigh individual elements of an RSSI scan when computing the global representation used for localization. Unlike the Set Transformer, this model uses a single learnable query vector to attend over the input sequence, yielding a weighted sum of embedded signal vectors. Each scan is encoded as an ordered sequence of vectors:
\begin{align}
s_i = [e_i \,\Vert\, r_i] \in \mathbb{R}^{d+1}, \quad i = 1, \dots, T
\end{align}

where $e_i$ is a learned embedding for BSSID $b_i$, and $r_i$ is the scalar RSSI value. These vectors are stacked into a matrix $S \in \mathbb{R}^{T \times (d+1)}$. A single learnable query vector $q \in \mathbb{R}^{d+1}$ is used to compute attention scores:
\begin{align}
\alpha_i = \frac{\exp(q^\top s_i)}{\sum_{j=1}^{T} \exp(q^\top s_j)}
\end{align}
These scores are used to form a weighted sum:
\begin{align}
z = \sum_{i=1}^{T} \alpha_i s_i
\end{align}

The resulting vector $z$ serves as a global summary of the scan, which is passed through an MLP head to predict the normalized 2D position:
\begin{align}
\hat{y} = \text{MLP}(z)
\end{align}

The model is trained using the same normalized coordinate targets and mean squared error objective described in Section~\ref{sec:training-objective}. The attention weights $\alpha_i$ are dynamically computed per scan, allowing the model to focus on the most informative access points.

This baseline serves to evaluate the utility of soft attention over RSSI sequences without the added architectural complexity of multi-head attention or permutation-invariant modules.

\subsection{Set Transformer Architecture (Main Model)}
\label{sec:set-transformer}

We adopt the Set Transformer architecture in \cite{lee2019set}, which is designed to operate on unordered sets and supports variable-sized input without imposing any sequential or spatial structure. This architecture is well-suited to our formulation of RSSI scans as permutation-invariant sets of $(\text{BSSID}, \text{RSSI})$ pairs.

The model begins by applying multiple \textit{Set Attention Blocks} (SAB), which use multi-head self-attention to model interactions among the elements of the input set $S = \{s_1, s_2, \dots, s_n\}$. Each SAB consists of a multi-head attention layer followed by a feedforward block with residual connections and layer normalization. These blocks allow the model to capture higher-order relationships between access points, including co-occurrence patterns and signal dependencies.

Following the attention blocks, a \textit{Pooling by Multihead Attention} (PMA) module aggregates the set into a fixed-size representation by attending to the entire set from a small set of learned seed vectors. In our implementation, we use a single seed vector to produce a global summary of the RSSI scan. This pooled representation is then passed through a fully-connected head to regress the 2D position output $(x, y)$.

Formally, the model defines a function $f_\theta: \mathcal{S} \rightarrow \mathbb{R}^2$, where $\mathcal{S}$ is the space of variable-length sets of encoded input vectors $s_i = [e_i \,\Vert\, r_i]$. The function $f_\theta$ is trained end-to-end using a regression loss described in Section~\ref{sec:training-objective}.

The Set Transformer’s ability to process unordered, variable-length input sets is essential for handling real-world RSSI scans. It enables the model to capture spatial signal structure while remaining robust to missing access points, variation in scan cardinality, and arbitrary ordering of detections.

\subsection{Training Objective and Loss Function}
\label{sec:training-objective}

The model is trained to regress the 2D position $(x, y)$ corresponding to each input RSSI scan. To ensure numerical stability and balanced optimization, we normalize the ground-truth coordinates using the empirical mean and standard deviation computed over the training set:
\begin{align}
x_{\text{norm}} = \frac{x - \mu_x}{\sigma_x}, \quad y_{\text{norm}} = \frac{y - \mu_y}{\sigma_y}
\end{align}
The model predicts normalized coordinates $(\hat{x}_{\text{norm}}, \hat{y}_{\text{norm}})$, which are rescaled at evaluation time using the inverse transform. We optimize model parameters by minimizing the mean squared error (MSE) between the predicted and true normalized positions:
\begin{align}
\mathcal{L}(\theta) = \frac{1}{N} \sum_{i=1}^N \left\| f_\theta(S_i) - (x_i^{\text{norm}}, y_i^{\text{norm}}) \right\|_2^2
\end{align}

where $f_\theta$ denotes the full Set Transformer model mapping RSSI input sets to position estimates, as described in Section~\ref{sec:set-transformer}. Training is conducted using the Adam optimizer with default hyperparameters. Each model is trained for a fixed number of epochs using early stopping based on validation set performance.

\subsection{Multi-Task Learning For Different Buildings, and Multiple Floors}
\label{sec:multi-task}

In the multi-task variant of our model, we introduce an auxiliary classification loss to predict the domain tag $z$ alongside the main regression task. Let $\hat{z}_i$ denote the predicted logits over domain classes for input $S_i$, and $z_i$ be the true class label. The total loss becomes:
\begin{align}
\mathcal{L}_{\text{total}}(\theta) = \mathcal{L}_{\text{reg}}(\theta) + \lambda \cdot \mathcal{L}_{\text{class}}(\theta)
\end{align}

where $\mathcal{L}_{\text{reg}}$ is the mean squared error loss defined above, $\mathcal{L}_{\text{class}}$ is the cross-entropy loss over domain labels, and $\lambda$ is a scalar weighting coefficient that balances the regression and classification terms. In our experiments, we use $\lambda = 1$.

Each of these scenarios tests the model’s ability to infer spatial position from signal patterns without relying on memorized access point identities. Furthermore, many BSSIDs in $\mathcal{D}_{\text{test}}$ may not be present in any training set (especially across buildings that are far apart). During inference, these unseen BSSIDs are assigned randomly initialized embeddings sampled from the same initialization distribution as those in training. The model must therefore generalize from learned structural patterns in RSSI space rather than relying on specific BSSID identity.

\section{Data Collection}
\label{sec:data-collection}

We now outline the data collection procedure starting with environmental setup, and field collection protocol.

\subsection{Environment Setup and Hardware}

All data was collected by scanning RSSI signals for pre-planned locations on a University campus. During collection, each scan would log RSSI values, BSSID identifiers, and timestamps. An iPad was used in parallel to mark up the path geometry, providing visual feedback to align the physical trajectory with later map annotation. To obtain ground-truth coordinates, post-hoc mapping was performed using satellite-based mapping software. This allowed for precise annotation of each step in Universal Transverse Mercator (UTM) coordinates. Domain floor plans were also acquired to aid in cross-verification of path alignments.

\subsection{Field Collection Protocol}

Data was collected at a university campus consisting of over 50 buildings, and 1.7 million square feet. Out of this, 6 buildings were selected some which contained only 1 floor, while others contained multiple floors. Collection involved scanning the Wi-Fi APs in each building, with some buildings requiring scans of multiple floors. The campus environments included buildings from typical university departments: Math Science, Earth Science, Engineering, Science A, Science B, Kinesiology, and Social Sciences, among others. Data collection was iterative spread over several sessions.

\subsection{Post-Processing}

Post-processing routines were applied to convert raw scan data into a standardized, model-ready format. Each scan was initially stored as nested JSON structures containing metadata and per-signal entries. Scripts were developed to parse these structures, extract relevant fields (such as RSSI, and BSSID), and flatten them into tabular format. Additional integrity checks were performed to handle malformed entries (bad data), remove duplicates, and ensure temporal consistency of sequential scans. These processed files were saved as CSVs with one row per scan, each representing a discrete spatial position. To support further transformations, we introduced a modular tagging system. A centralized JSON mapping file maintained metadata about the buildings and floors on campus in a scalable way to allow for future expansions of this work to other indoor environments.

\subsection{Data Preparation}

Following post-processing, each scan was transformed into a fixed-length numerical feature vector. This included mapping all observed BSSIDs to consistent column indices across datasets. The resulting feature vectors were normalized using min-max bounds computed per-building and saved separately for reproducibility (and segmented quality checking).

Coordinate transformations were applied to convert manually annotated positions into globally consistent UTM coordinates. Domain and floor tags were appended to each example to enable domain-specific evaluation protocols. The final training and evaluation splits were constructed based on the experimental setup described in Section~\ref{sec:experiments}. All prepared examples were stored in a flattened CSV format with aligned feature-label pairs ready for experimentation and modeling.

\subsection{Dataset Overview}
\label{sec:dataset-overview}

Table ~\ref{tab:building-data} below summarizes the final data set for this study after the data post-processing, and data preparation steps. Some of the building names are anonymized for privacy reasons, using names like Building 1 and Building 6 in place of the real name. In total, there were 3606 examples counting the first floors of each building in Table ~\ref{tab:building-data}. The multi-floor experiments outlined in ~\ref{sec:experiment-setups} are for Building 1 which contained 3 floors. The total examples in the data set for Building 1 only along with all three floors consisted of 1929 scans (so 1929 examples in total).

\begin{table}[h]
\centering
\small
\caption{Overview of buildings used for data collection, and number of examples obtained}
\vspace{1em}
\begin{tabular}{lccr}
\toprule
\textbf{Building} & \textbf{Floors} & \textbf{Area (m$^2$)} & \textbf{Total Scans (first floors)} \\
\midrule
Building 1 & 3 & 4,961.331 & 647 \\
Math Science & 3 & 1,790.950 & 770 \\
Earth Science & 1 & 3,031.455 & 376 \\
Science B & 2 & 4,680.266 & 299 \\
Science A & 2 & 4,615.369 & 1,065 \\
Building 6 & 2 & 2,519.332 & 449 \\
\midrule
\textbf{Total} & \textbf{13} & \textbf{21,598.703} & \textbf{3,606} \\
\bottomrule
\end{tabular}
\label{tab:building-data}
\end{table}

\section{Experiments}
\label{sec:experiments}

We now outline the experimental set up of our work.

\subsection{Benchmarking}
\label{sec:benchmarking}

We benchmark the Set Transformer against several standard neural architectures widely used in structured and unstructured input modeling. These baselines represent different paradigms: fixed-length vector modeling, sequential processing, and attention-based summarization. The purpose here is to evaluate the impact of architectural design on indoor localization accuracy.

The baseline models were introduced in ~\ref{sec:methods}. These models consist of a Multilayer Perceptron (MLP) that maps fixed-length RSSI vectors to coordinates using a fixed BSSID index, a vanilla Recurrent Neural Network (RNN) and a Long Short-Term Memory (LSTM) network that process sorted $(\text{BSSID}, \text{RSSI})$ sequences, a lightweight attention model using a single query vector, and the Set Transformer which treats scans as unordered sets. These baselines were chosen for their widespread use in modeling vector, sequential, or set-based data, offering a representative benchmark. To ensure a fair comparison, all models were trained with matched hyperparameters and capacity constraints: identical learning rate ($10^{-3}$), batch size (32), fixed number of training epochs (50), Adam optimizer, no dropout, and minimal tuning. Architectural differences aside, every model received the same preprocessed data, enabling direct performance comparison.

\subsection{Experimental Configurations}
\label{sec:experiment-setups}

We evaluate all models under three supervised tasks of increasing spatial and structural complexity. These are E1 (for Experiment 1), E2, and E3 below.

\paragraph{E1. Single Building, Single Floor:} This experiment has the simplest set up consisting of only one floor of a single building on campus. Each model is trained using the collected data from this floor using supervised learning: the train and test data are assumed to be from the same distribution. This setting minimizes architectural variation and serves as a controlled benchmark for signal-to-position learning capacity: if models fail at this stage, they will likely fail in more complicated set ups (E2, and E3, but also future works that extend to different domains).

\paragraph{E2. Multiple Buildings, First Floors only:} In this experiment, we extend modeling to the data collected on the first floor of multiple buildings. This allows for a wider variety of BSSIDs since buildings further apart are less likely to share the same access points. Again, this is a supervised problem where the training, and test sets assume the same distribution, and the held out test set reasonably contains examples from all the buildings in E2. This setting evaluates horizontal generalization across buildings with different layouts and infrastructure.

\paragraph{E3. Single Building, Multiple Floors:} Lastly, in this experiment we consider a single building with multiple floors. The training and test sets again assume the same distribution with the test set reasonably containing examples from each floor. This experiment evaluates vertical generalization and the model’s ability to disambiguate inter-floor signal overlap.

\subsection{Evaluation Protocol}
\label{sec:evaluation-protocol}

The task is 2D coordinate regression: given a single RSSI scan, each model predicts a point $(x, y)$ in physical space. Performance is evaluated using the \textbf{mean Euclidean distance} between predicted and ground-truth coordinates on the test set. This is computed as:
\begin{align}
\text{Error} = \frac{1}{n} \sum_{i=1}^{n} \sqrt{(\hat{x}_i - x_i)^2 + (\hat{y}_i - y_i)^2}
\end{align}

We also report the \textbf{standard deviation} of this per-sample distance to capture prediction variability. Coordinates are normalized using training set statistics for each experiment and denormalized at inference time to compute errors in meters. For qualitative assessment, we include 2D and 3D visualizations comparing predicted and ground-truth locations across different environments.

\section{Results}

Table ~\ref{tab:exp-results} summarizes the results across E1 - E3 for each of the 4 baseline models, and the main Set Transformer model.

\begin{table}[h]
\centering
\small
\caption{Summary of experimental results across tasks E1–E3. All values are mean ± standard deviation of 2D test error in meters.}
\vspace{1em}
\begin{tabular}{lcc}
\toprule
\textbf{Experiment} & \textbf{Model} & \textbf{Test error (m)} \\
\midrule
\multirow{5}{*}{E1: Single building, single floor}
 & MLP & 7.54 ± 4.75 \\
 & RNN & 7.92 ± 8.05 \\
 & \textbf{LSTM} & \textbf{2.23 ± 2.19} \\
 & Attention & 5.77 ± 3.10 \\
 & Set Transformer & 3.82 ± 2.35 \\
\midrule
\multirow{5}{*}{E2: Multiple buildings, first floor only}
 & MLP & 49.12 ± 19.48 \\
 & RNN & 20.51 ± 16.86 \\
 & \textbf{LSTM} & \textbf{3.13 ± 3.23} \\
 & Attention & 13.04 ± 8.77 \\
 & Set Transformer & 6.30 ± 5.14 \\
\midrule
\multirow{5}{*}{E3: Single building, multiple floors}
 & MLP & 15.68 ± 6.29 \\
 & RNN & 29.09 ± 14.49 \\
 & \textbf{LSTM} & \textbf{2.44 ± 4.18} \\
 & Attention & 7.20 ± 5.74 \\
 & Set Transformer & 3.53 ± 4.28 \\
\bottomrule
\end{tabular}
\label{tab:exp-results}
\end{table}

Across all three experiments, the LSTM consistently achieved the lowest mean localization error, outperforming all other models. The Set Transformer ranked second in every setting, demonstrating stable performance across varying spatial conditions. A substantial drop in accuracy was observed when moving from E1 to E2, where scans were drawn from multiple buildings with different access point distributions. Similarly, performance declined from E1 to E3 for most models, though the LSTM and Set Transformer remained relatively robust across multiple floors.

\begin{figure}[p] 
  \centering
  \includegraphics[width=\textwidth,height=\textheight,keepaspectratio]{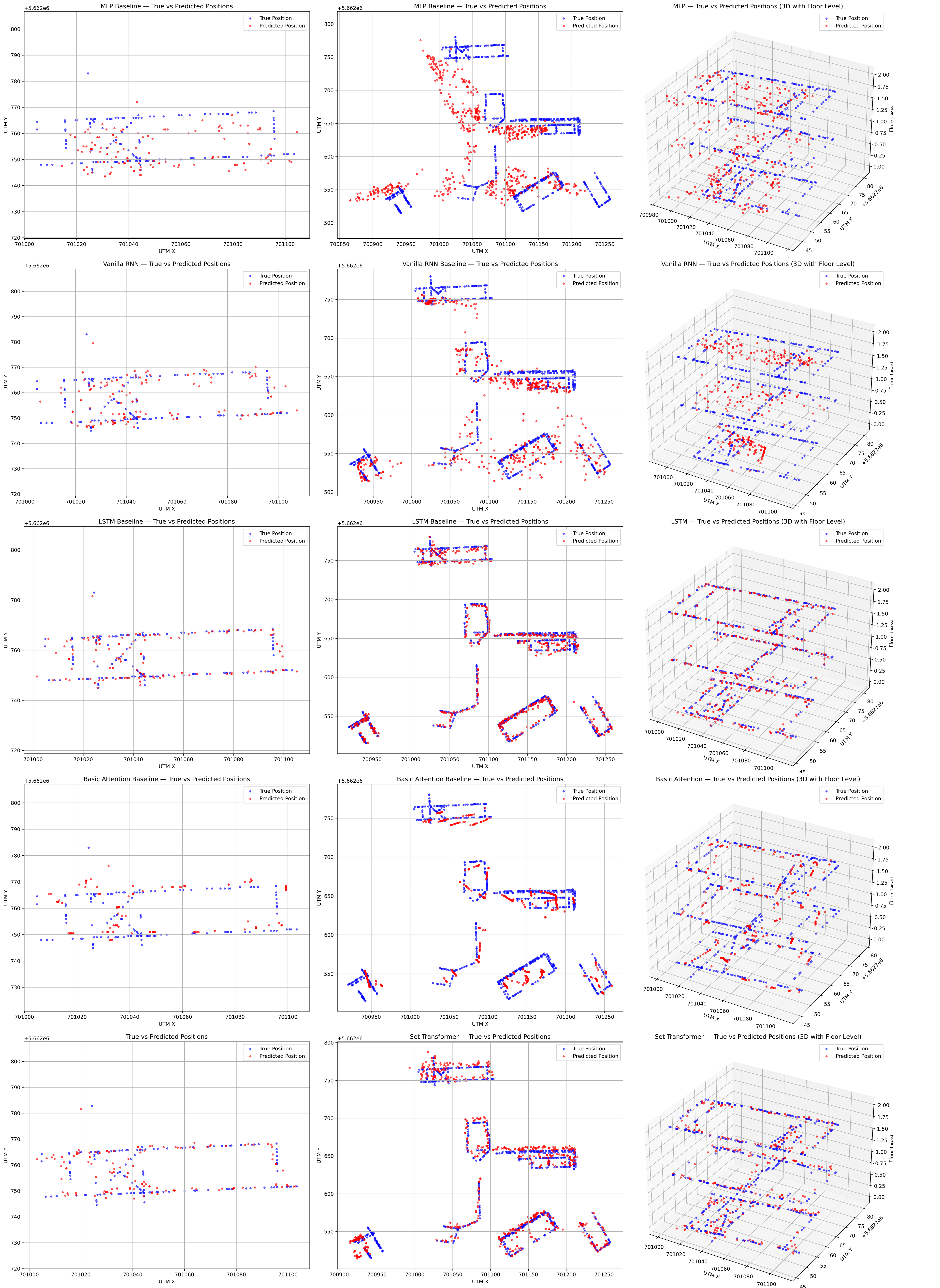}
  \caption{Comparison of model predictions across experiments (columns: E1, E2, E3) and model types (rows: MLP, RNN, LSTM, Attention, Set Transformer).}
  \label{fig:full_panel}
\end{figure}

Figure~\ref{fig:full_panel} visualizes the predicted and true locations across all models and tasks. In E1 (left column), most models follow the corridor-like structure of the environment, but only the LSTM and Set Transformer maintain consistently accurate alignment with the true trajectories. The MLP performs poorly, producing predictions that spill into rooms and walls not traversed during data collection.

In E2 (middle column), prediction quality degrades significantly for all models except the LSTM and Set Transformer. The MLP and RNN baselines scatter predictions throughout non-navigable space, including outside the buildings themselves. This reflects their difficulty generalizing across differing Wi-Fi infrastructures and layouts. The attention model captures some structure but fails to localize consistently.

In E3 (right column), which includes multiple floors, only the LSTM and Set Transformer preserve the layered structure of the building in 3D space. The other models struggle to maintain vertical separation, with the MLP and RNN showing substantial misalignment between floors and frequent cross-floor leakage. The Set Transformer’s ability to infer spatial grouping without relying on ordering proves effective in maintaining floor-specific predictions.

Overall, the plots confirm the quantitative findings: LSTM achieves the highest spatial fidelity, while the Set Transformer offers a close second with more stable predictions than the simpler baselines. We discuss scenarios where the advantages of the Set Transformer may overcome the LSTM and other models.

\section{Discussion}

The results across Experiments E1–E3 reveal consistent trends in model behavior and highlight both the promise and limitations of applying set-based neural architectures to the indoor localization problem.

\subsection{Performance Across Tasks}

In all experimental settings, the LSTM model achieved the lowest average test error, with particularly strong results in the single-floor (E1) and multi-building (E2) settings. Its sequential processing capabilities - despite the input not being explicitly temporal - enabled it to learn consistent spatial features across scans. This effect is especially visible in the visualizations, where LSTM predictions align closely with the hallway structures and true paths, even when the geometry varies across buildings.

The Set Transformer also performed strongly across all three experiments, showing robustness to varying input set sizes and spatial permutations. Its permutation-invariant architecture appears to confer a generalization advantage, particularly on E3 (multi-floor). While it slightly underperformed LSTM numerically, its predictions were comparably structured in 2D and 3D plots.

In contrast, the MLP baseline failed to generalize in all but the simplest setting (E1), with large errors and erratic predictions on multi-building and multi-floor tasks (even in the simplest setting however, the errors were still high and not acceptable based on the visualization). The RNN improved upon MLP but lacked the capacity to model longer-range dependencies. Attention-based models demonstrated moderate success (especially in E2) but often produced vertically misaligned or diffused predictions, likely due to sensitivity to signal sparsity and noise.

Visual inspection of the 2D and 3D plots confirms the quantitative findings. In E1, only the LSTM and Set Transformer produced tight prediction paths. In E2, both models preserved building structures, with LSTM recovering path layouts across diverse floor plans. In E3, the 3D visualizations show clean floor separation and alignment for these two models, whereas others exhibited either floor mixing (RNN, MLP) or vertical drift (Attention).

\subsection{Embedding-Based Generalization and Permutation-Invariance}

Our results further validate the effectiveness of embedding-based representations in supporting generalization across diverse signal environments. By converting categorical BSSID identifiers into dense, learnable vectors, the model can extract structural patterns from RSSI sets without relying on memorized access point identities. This representation is critical for permutation-invariant architectures like the Set Transformer, which must reason over unordered sets of signals.

The superiority of the Set Transformer and LSTM across multiple settings—particularly in E2 (first floor of multiple buildings) and E3 (multiple floors of a single building)—suggests that the model successfully leverages learned embeddings to identify latent structure such as spatial proximity and co-occurrence patterns between access points. Notably, the performance gap between MLP/RNN baselines and embedding-based attention models becomes more pronounced in environments with higher BSSID variability, reinforcing the role of learned signal relationships in supporting generalization.

Moreover, our approach permits graceful handling of unseen BSSIDs at inference time. Since each BSSID is mapped to an embedding during training, new identifiers encountered in the test set can be assigned randomly initialized vectors without requiring architectural changes or retraining. Despite this, the Set Transformer maintains competitive accuracy, indicating that the model has learned transferable signal-to-location mappings that go beyond simple identity matching. This is especially important for scalability in real-world deployments, where AP inventories vary across buildings and over time.

\subsection{When Permutation-Invariance May Offer Advantages}

While the LSTM slightly outperforms the Set Transformer in our experiments, we hypothesize that permutation-invariant architectures may offer superior performance in several practical scenarios. First, when the ordering of RSSI measurements is arbitrary or unstable due to hardware inconsistencies, scanning latencies, or platform-specific APIs, a set-based model can better preserve the semantic integrity of the input. Sequence models like LSTMs may inadvertently learn spurious patterns tied to ordering, reducing their robustness in deployment.

Second, in deployments where the access point infrastructure is fixed but large (such as multi-floor campuses or corporate buildings) the signal scan may still yield variable subsets of BSSIDs due to occlusion, interference, or partial coverage. In such cases, the presence or absence of specific APs across scans introduces sparsity and inconsistency in the input. Permutation-invariant models can handle this variability more naturally, since their performance does not depend on the presence of a fixed number or order of AP features.

Finally, Set Transformers may be better suited for multi-modal signal fusion: combining BLE, Wi-Fi, and other sensor modalities, where each modality provides a set of observations with no inherent ordering. The ability to flexibly integrate such data while reasoning over signal-level relationships could prove beneficial in richer localization contexts where sequence-based models are harder to apply consistently.

In summary, other future works could involve testing these hypotheses aimed to further draw out the robustness of set-based neural architectures over high-performing baselines like the LSTM.

\subsection{Key Takeaways}

Our results highlight several important insights. First, sequential models such as LSTMs prove remarkably effective for RSSI-based localization, outperforming all other baselines despite the unordered nature of the input data. This suggests that their inductive biases, particularly their ability to integrate contextual dependencies, compensate for the lack of input structure. Set Transformers also deliver competitive performance, consistently ranking second across experiments. Their permutation-invariant architecture and ability to model access point relationships at the set level make them a promising candidate for scalable, domain-agnostic localization. In contrast, simpler architectures such as MLPs and vanilla RNNs struggle to generalize beyond narrow training distributions, performing poorly in settings with cross-building variation or vertical complexity. Finally, visualization plays a critical role in model assessment: spatial scatter plots reveal nuanced patterns of misalignment and degradation that summary metrics such as mean error may obscure. These findings inform future work on domain adaptation and generalization, where models must make predictions to unseen spaces while preserving fine-grained spatial accuracy.

\subsection{Limitations}

Despite the promising results, this study has several limitations that constrain the generalizability and applicability of its findings. First, all data were collected under a controlled protocol involving straight-line paths on building hallways. While this consistency reduced measurement noise and facilitated label alignment, it departs from realistic user behavior in deployed scenarios, potentially overestimating model robustness. Second, the experimental scope is limited to indoor university buildings with relatively homogeneous architectural layouts and minimal radio frequency interference. As a result, the findings may not transfer directly to more complex environments such as shopping malls, airports, warehouses, or hospitals.

Additionally, the models rely solely on received signal strength (RSSI) measurements as input. Although RSSI offers widespread availability and low-cost integration, it is inherently noisy and prone to variability due to multipath propagation and environmental changes. The absence of complementary modalities, such as inertial measurements or barometric pressure, limits resilience under challenging conditions. Further, each scan is treated independently, with no temporal modeling or sensor fusion. This inhibits the use of continuity cues that could aid in filtering transient errors.

Architectural choices were also not extensively ablated. While multiple baselines were compared, we did not explore the sensitivity of the Set Transformer to its internal hyperparameters (such as attention heads, or embedding dimensions), nor did we examine whether model depth or regularization strategies would yield improvements. The focus of this work was just to ensure that the models were at the very least, comparable, which we accomplished by ensuring they were all trained under the same conditions. In future, we can conduct both ablation studies, and further parameter tuning on each model, followed by another comparison. Lastly, no latency or runtime benchmarks were conducted. Although models were evaluated on accuracy, their computational efficiency, memory requirements, and suitability for real-time deployment on mobile hardware remain untested.

Addressing these limitations is a priority for future work, particularly in transitioning from research settings to robust, real-time localization systems deployed in heterogeneous real-world environments.

\section{Conclusion}

In this work, we investigated the application of permutation-invariant neural architectures for indoor localization using RSSI-based signal data. We introduced a Set Transformer model designed to handle unordered signal scans and benchmarked its performance against a suite of baseline architectures, including the MLP, RNN, LSTM, and attention model.

We evaluated these models on three representative localization scenarios: (E1) single building, single floor, (E2) multiple buildings, but containing the first floor only, and (E3) single building with multiple floors. Our results demonstrate that while traditional sequence models such as LSTMs perform competitively in most scenarios, the Set Transformer offers a compelling trade-off between accuracy and architectural generality particularly when input ordering is unstable or arbitrary.

Despite not outperforming LSTM in all cases, the Set Transformer remains promising due to its inductive bias toward unordered data and capacity to operate without handcrafted feature sorting. This paper lays the foundation for more generalizable and robust neural approaches to indoor localization using signal-based data.

Future work will explore domain adaptation and generalization, large-scale deployment, and integration with mobile systems to enable real-time inference on mobile systems.

\bibliographystyle{unsrt}
\bibliography{references}


%

\end{document}